\documentclass[article]{article}

\usepackage[english]{babel}
\usepackage{indentfirst}

\usepackage{amsmath}
\usepackage{graphicx}
\usepackage[colorinlistoftodos]{todonotes}
\usepackage[colorlinks=true, allcolors=blue]{hyperref}
\usepackage{diagbox}
\usepackage{slashbox}
\usepackage{multirow}

\usepackage{appendix}

\begin{document}

\title{Benchmarks of ResNet Architecture for Atrial Fibrillation Classification}
\author{
  Roman Khudorozhkov\\
  {\tt\small r.khudorozhkov@analysiscenter.org}
  \and
  Dmitry Podvyaznikov\\
  {\tt\small d.podvyaznikov@analysiscenter.org}
}

\maketitle

\begin{abstract}
In this work we apply variations of ResNet architecture to the task of atrial fibrillation classification. Variations differ in number of filter after first convolution, ResNet block layout, number of filters in block convolutions and number of ResNet blocks between downsampling operations.
We have found a range of model size in which models with quite different configurations show similar performance. It is likely that overall number of parameters plays dominant role in model performance. However, configuration parameters like layout have values that constantly lead to better results, which allows to suggest that these parameters should be defined and fixed in the first place, while others may be varied in a reasonable range to satisfy any existing constraints. 
\end{abstract}

\section{Introduction}
In the past years neural networks have surpassed classic approaches in a number of tasks and achieved state-of-the-art results. The main domains of application were image classification, object detection / segmentation and speech recognition. This became possible by virtue of two factors: development of hardware and emergence of large open-sourced datasets, such as Imagenet.

The medical domain has been difficult to lever with neural networks for quite a long time, mostly due to the lack of large quality datasets. The tables have turned a few years ago, and since then a number of large datasets has been released, mostly in a form of competition. This allowed many engineers to contribute in solving medical problems, such as knee osteoarthritis diagnosis \cite{Knee} and bone age assessment~\cite{Bone}. But in some tasks, such as ECG classification, there are few large open datasets, and most of the publications are either supported by private data, or make use of small datasets, which may not be representative and unbiased \cite{Egor}.

In this paper we apply variations of convolutional neural network architecture - ResNet \cite{Resnet} - to the task of atrial fibrillation classification to obtain benchmarks and get intuition behind performance of those variations.

\section{Dataset}
In this work we use publicly available training part of the PhysioNet Computing in Cardiology Challenge 2017 dataset \cite{PN2017,Pnet}. It contains 8528 signals, with length from 9 to 61 seconds. The signals are single-lead recordings sampled as 300 Hz and band pass filtered by portable device. The objective of the challenge was to differentiate ECG signals in four categories: (A) signals with atrial fibrillation, (N) signals with normal rhythm, signals with other abnormal rhythms (O) and signals too noisy to be classified ($\sim$).  According to the latest version of labels, there are 758 records with atrial fibrillation, 5076 records with normal rhythm, 2415 records with other abnormal rhythms and 279 noisy records.

\section{Methods}

\subsection{Data preparation}
In this work we removed noisy signals ($\sim$) and consolidated classes (N) and (O), thus obtaining dataset for binary classification: atrial fibrillation (A) vs non-atrial fibrillation (NO) with 758 and 7491 signals respectively. Resulting dataset was randomly split into 80\% training and 20\% validation sets.

Due to specifics of the heart monitoring device, many signals in the dataset have their R-peaks directed downwards. Those signal were flipped during the preprocessing phase.

For training we used random crops of approximately 10 second long. Under the assumption that heart rhythm class does not change during the whole recording, those crops were labeled according to the label of the original recording. To eliminate class imbalance, we sampled three times more crops from recordings with atrial fibrillation compared to other recordings.
Also, as an augmentation technique we adopted signal resampling to a new randomly chosen sampling rate, which allowed to represent variability of heart rates.

\subsection{Model variations}
In this work we test performance of ResNet-like architectures.
For ResNet-like models we vary following parameters: number of filters after the first convolution operation; number of ResNet blocks in the network and their layout; number of filters in each block.

To describe model configuration we will use the following form:
\begin{align*}
	\{input filters; \; layout; \; filters; \; blocks\},
\end{align*}
e.g.
\begin{align*}
	\{32; \; cna; \; 
    [4,\; 8,\; 16,\; 32,\; 64,\; 128,\; 256]; \;
    [1,\; 1,\; 1,\; 1,\; 1,\; 1,\; 1]\}.
\end{align*}
To see how elements of this configuration correspond to ResNet architecture, see Table~\ref{tab:resarchitecture}.
Layout is represented as a consequence of layers, described by the first letter. Thus, \{cna\} is a consequence of (c)onvolution, (n)ormalization and (a)ctivation layers. We use 1D convolutions, batch normalization~\cite{BN} and ReLu~\cite{relu} as activation function. Also, we build skip connections after each ResNet block, which is represented by 
\begin{align*}
\bigg[ [3, \; filters] \times  layout \bigg].
\end{align*}
A number of ResNet blocks form a group, at the end of which we downsample the signal by a factor of two.

\begin{table}[th!]
\begin{center}
\def\arraystretch{1.1}
\begin{tabular}{| c | c |}
\hline
			 & ResNet model
\\  \hline
Layer        & Block parameters
\\ \hline
Conv block	 & 7, \textit{input filters}, /2
\\ \hline
ResNet group 1 & \( \bigg[ \hspace{-0.4em} \begin{array}{l}
                                [3, \; filters_1] \times layout \\
                  \end{array} \bigg] \times blocks_1 \) 
\\ \hline
ResNet group 2 & \( \bigg[ \hspace{-0.4em} \begin{array}{l}
                                [3, \; filters_2] \times layout \\
                  \end{array} \bigg] \times blocks_2 \) 
\\ \hline
...            &  ...
\\ \hline
ResNet group n & \( \bigg[ \hspace{-0.4em} \begin{array}{l}
                                [3, \; filters_n] \times layout \\
                  \end{array} \bigg] \times blocks_n \) 
\\ \hline
Classification layer &
\\ \hline
\end{tabular}
\end{center}
\caption{Configuration parameters for ResNet architecture.}
\label{tab:resarchitecture}
\end{table}

In this work we test 28 configurations along with classic ResNet18 and ResNet34 architectures. Full list of tested models, trainable parameters and performances is presented in Appendix~\ref{tables}.

\subsection{Training}
All models were trained with Adam optimizer~\cite{Adam}. The batch size was set to 32 to avoid memory issues while training large networks. Each model variation was trained independently 5 times for 300 epochs with CardIO framework \cite{Cardio} available at \url{https://github.com/analysiscenter/cardio}. 

We repeat the training process five times to obtain a distribution of results rather than a point estimate, because result of a single run can be inaccurate due to high variability of training process.

\subsection{Inference}
To obtain a prediction for a signal we used following procedure: each signal was split into crops without overlapping; predictions for crops from the same signal were averaged and the resulting value compared with threshold equal to 0.5. 

It is important to notice that we did not apply any augmentation during testing phase.

\section{Results}
We use F1 score metric to compare performance of model variations described in Appendix~\ref{tables} on test set.
Also, we investigate how different configuration parameters affect performance as the overall number of parameters in the network change.

\subsection{Number of parameters in a model}
\begin{figure*}[!h]
\begin{center}
\includegraphics[width=\linewidth]{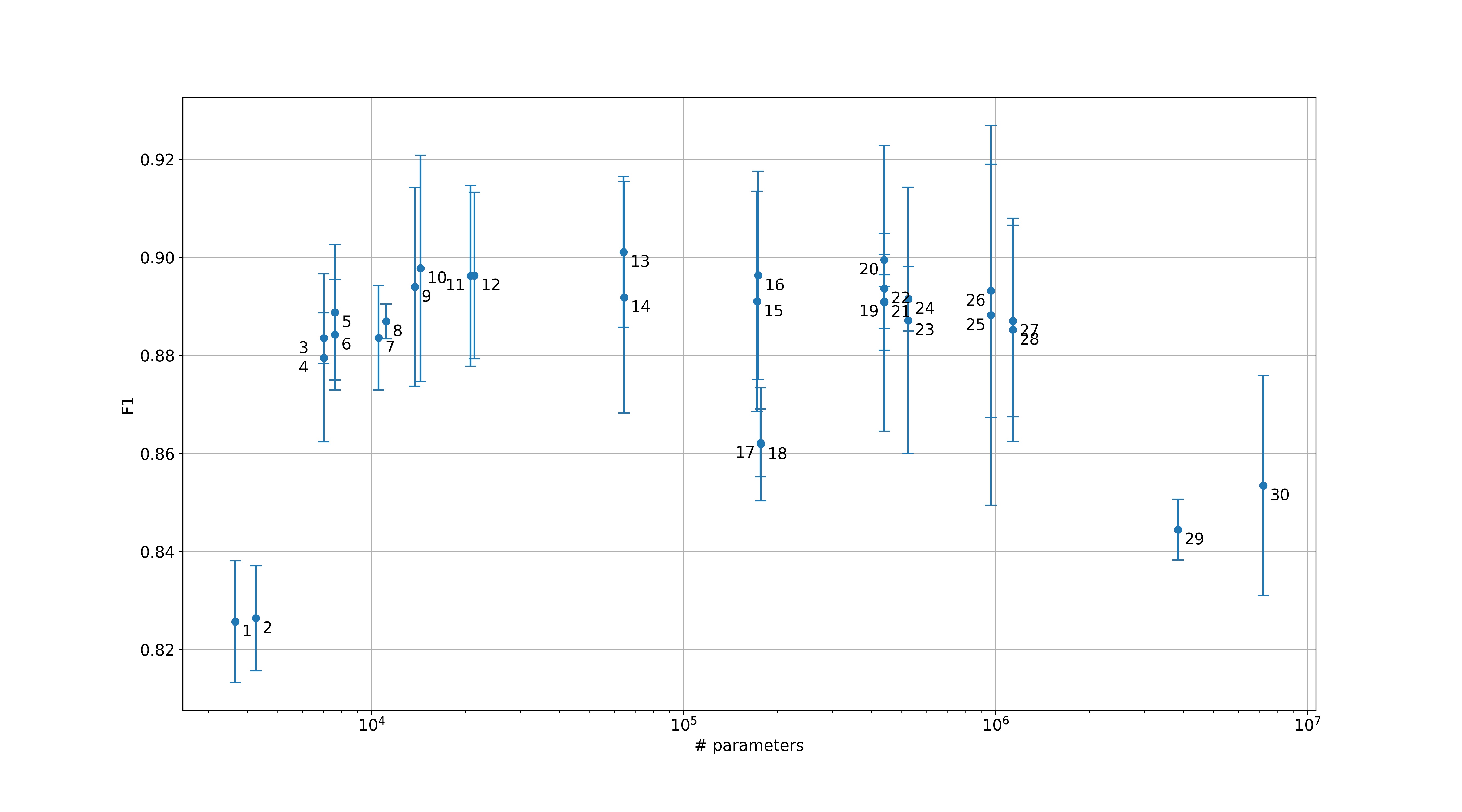}
\end{center}
\caption{F1 score metric of model performance and number of parameters in the model.}
\label{fig:parameters}
\end{figure*}

Fig.~\ref{fig:parameters} displays dependence of F1 metric of the model on the number of parameters in the model. The point number in the figure correspond to a model number in Appendix~\ref{tables}. In our experiments model size varied in a range between $10^3$ and $10^7$ parameters. It is clear that models with less than $10^4$ and more than $10^6$ parameters perform poorly, while models inside this range show the best results. However, overall number of parameters is not the only important hyperparameter: although configurations \{x; cna; [4, 8, 16, 32, 64, 128, 256]; [1, 1, 1, 1, 1, 1, 1]\} perform better than \{x; cna; [4, 4, 8, 8, 16, 16, 20]; [1, 1, 1, 1, 1, 1, 1]\}, they do not match configurations with same number of parameters and more reasonable configuration of blocks and filters.

In following subsections we will pay attention to each component of the configuration separately and try to understand their role in model performance.

\subsection{Input filters}

\begin{figure*}[!ht]
\begin{center}
\includegraphics[width=\linewidth]{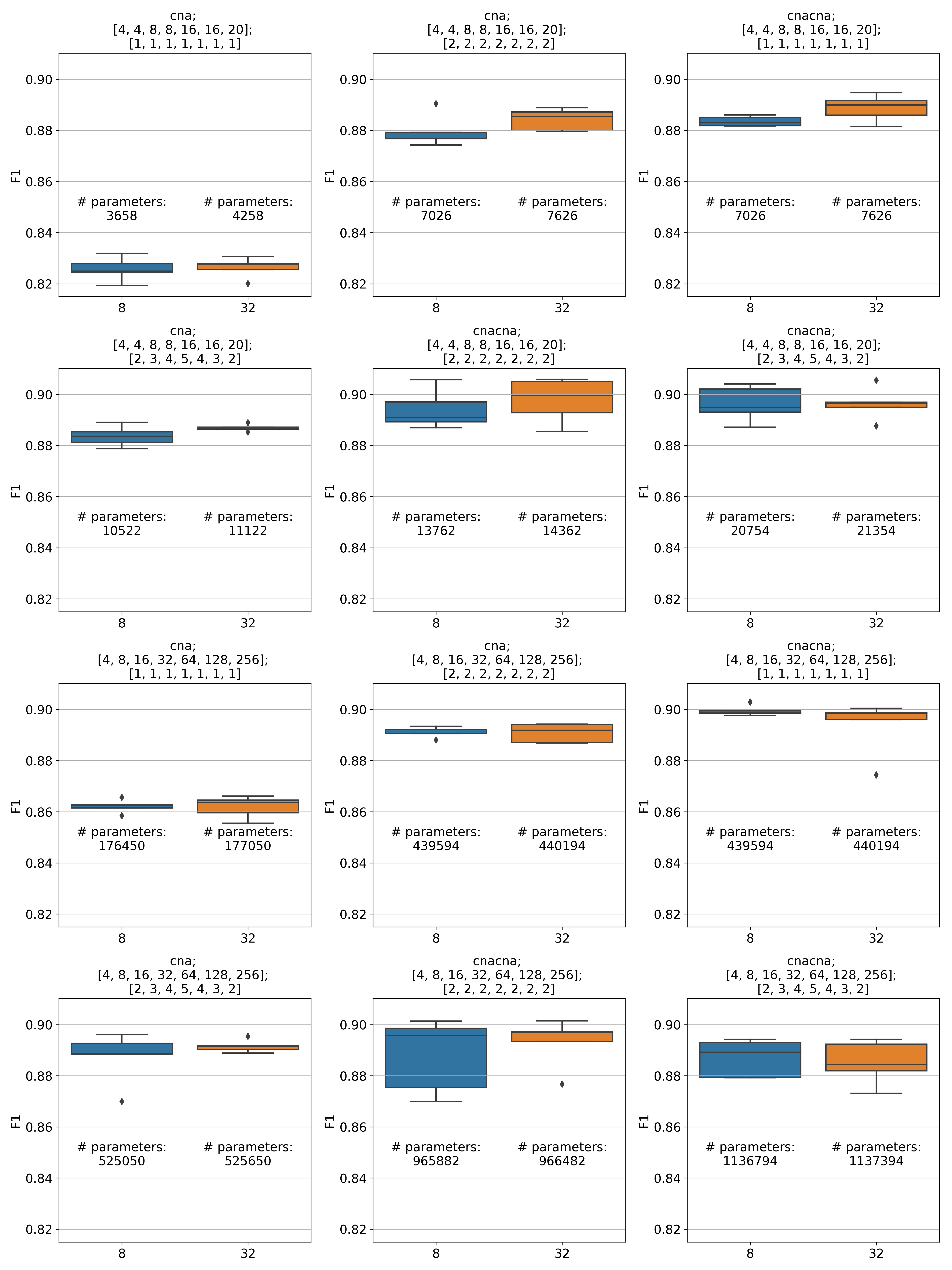}
\end{center}
\caption{F1 metric of model performance on the number of input filters.}
\label{fig:input_filters}
\end{figure*}

On the Fig.~\ref{fig:input_filters} you can see that number of filters in the first convolutional layer does not affect performance for a very small model, such as \{cna; [4, 4, 8, 8, 16, 16, 20]; [1, 1, 1, 1, 1, 1, 1]\}, which has less than 5 thousand parameters. As the number of parameters increase up to about $10^5$, greater number of input filters tend to become more useful for the model. As model size grows bigger, the difference between 8 and 32 input filters becomes insufficient or even in favor of smaller number of input filters.

It is likely that smaller networks highly depend on the first feature maps as they are limited with subsequent processing powers. As the network grows, it relies less on the early feature maps as it can extract needed information using convolutions in the later layers.

\subsection{Layout}
\begin{figure*}
\begin{center}
\includegraphics[width=\linewidth]{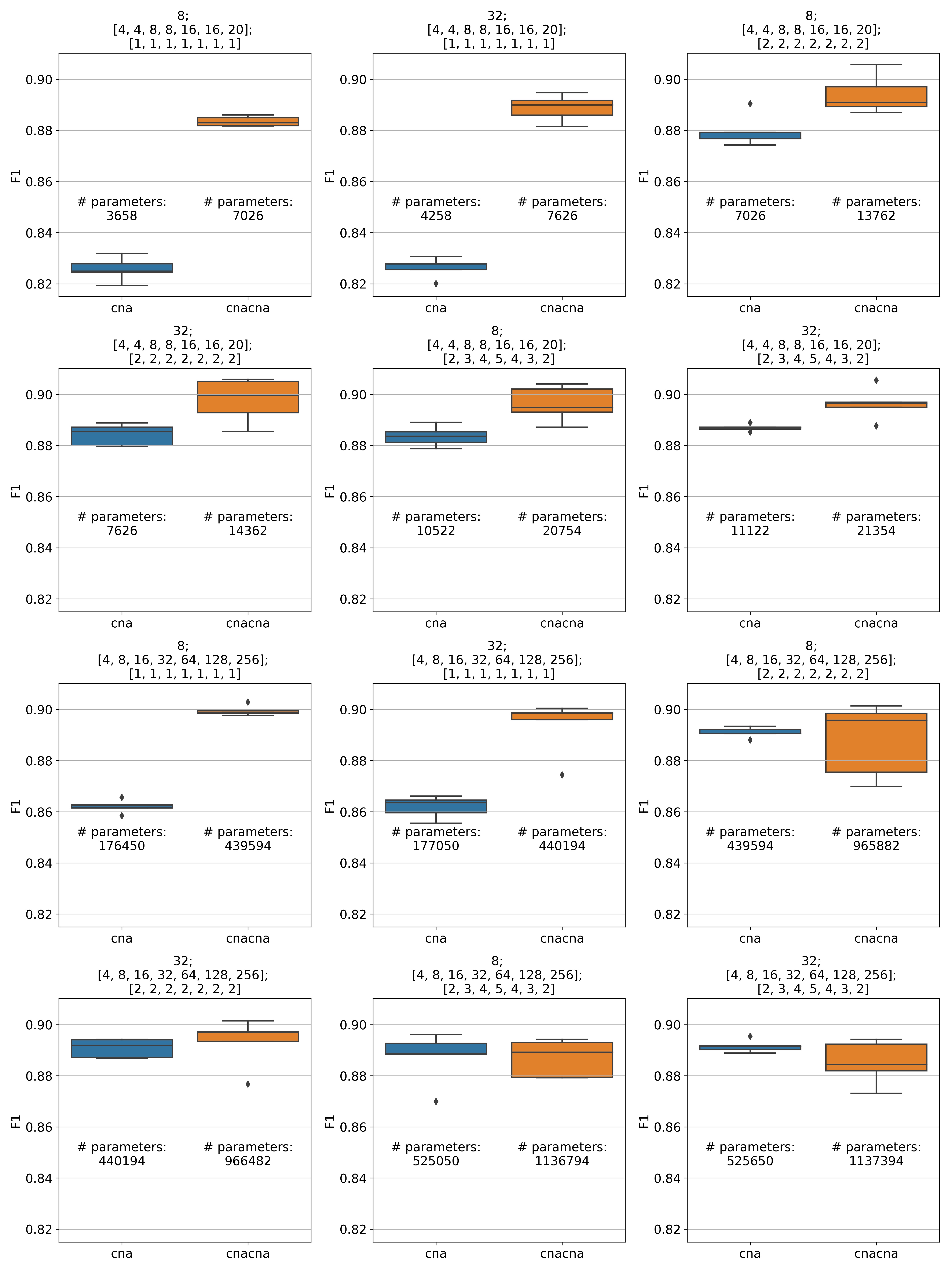}
\end{center}
\caption{F1 metric of model performance on layout.}
\label{fig:layout}
\end{figure*}

In Fig.~\ref{fig:layout} we compare performance of models with \{cna\} and \{cnacna\} layouts. As the figure shows, configurations with two convolutions in ResNet block significantly outperform those with only one convolution. Nevertheless, when models with \{cnacna\} layout become as big as $10^6$ parameters, they stagnate and even start to drop in performance, while models with \{cna\} layout start to express such behavior at about $4\times10^5$ parameters

\begin{figure*}
\begin{center}
\includegraphics[width=\linewidth]{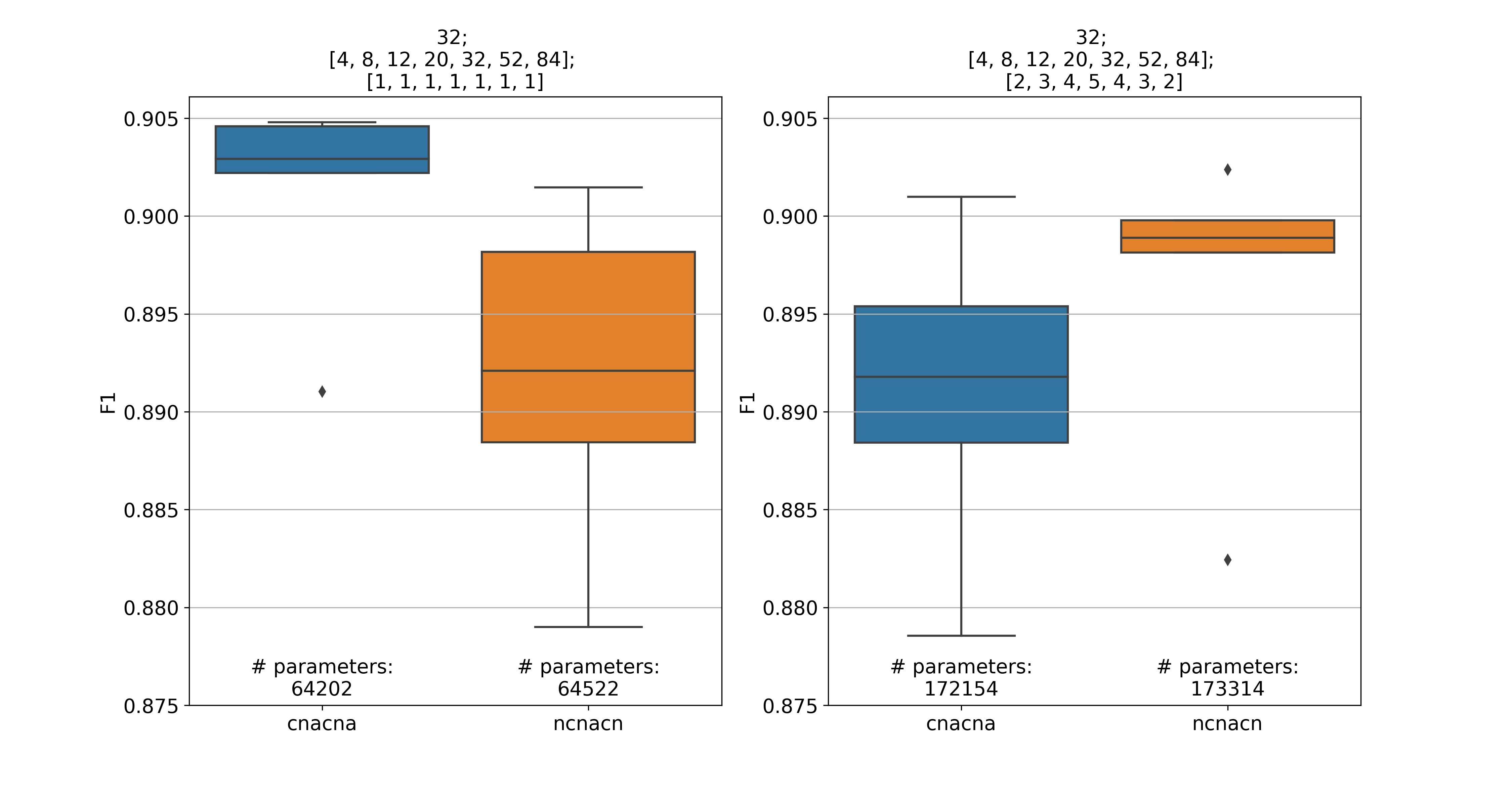}
\end{center}
\caption{F1 metric of model performance on layout.}
\label{fig:layout2}
\end{figure*}

Another figure, Fig.~\ref{fig:layout2}, shows comparison of \{cnacna\} layout and \{ncnacn\} layout, which was suggested in \cite{pyramid}. Configurations with different layouts have almost the same number of parameters so only layout makes the difference between them. Interestingly, \{cnacna\} performs better when number of blocks is small and overall number of parameters is about $64\times10^3$; \{ncnacn\} perform better as the number of blocks increases and the model has about $173\times10^3$ parameters. However, we need to conduct more experiments to prove this pattern.

\subsection{Filters}
\begin{figure*}
\begin{center}
\includegraphics[width=\linewidth]{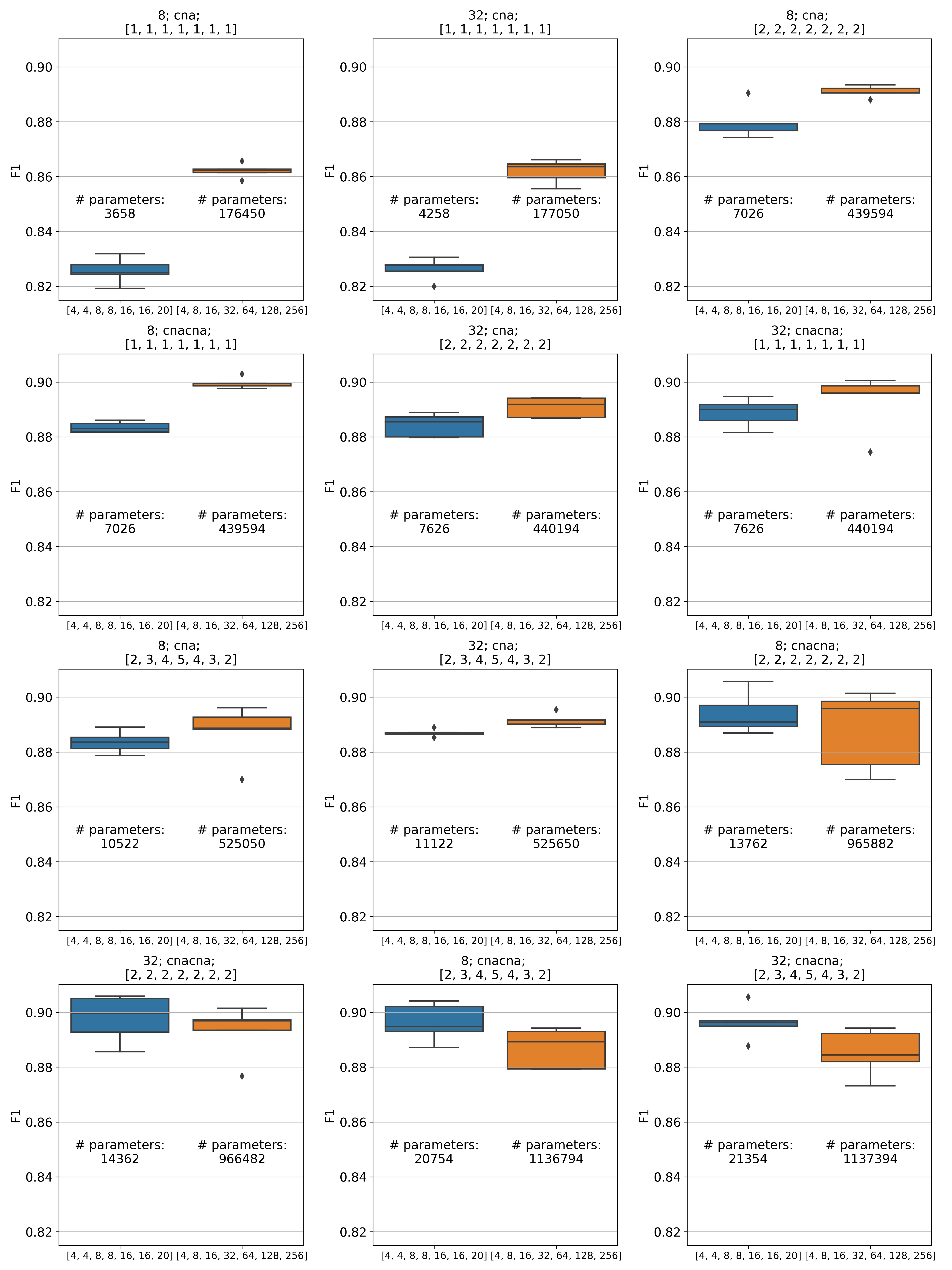}
\end{center}
\caption{F1 metric of model performance on the filters configuration.}
\label{fig:filters}
\end{figure*}

The Fig.~\ref{fig:filters} shows that more filters generally lead to better performance. This statement holds true for models of both \{cna\} and \{cnacna\} layouts and across all tested configurations of blocks. However, big models with more than $9\times10^5$ parameters tend to perform worse than models with small filter sizes. 

This effect is most likely to be a result of too big capacity of the model, which makes it difficult to converge to a better optima in a very complex parameter space.

\subsection{Blocks}
\begin{figure*}
\begin{center}
\includegraphics[width=\linewidth]{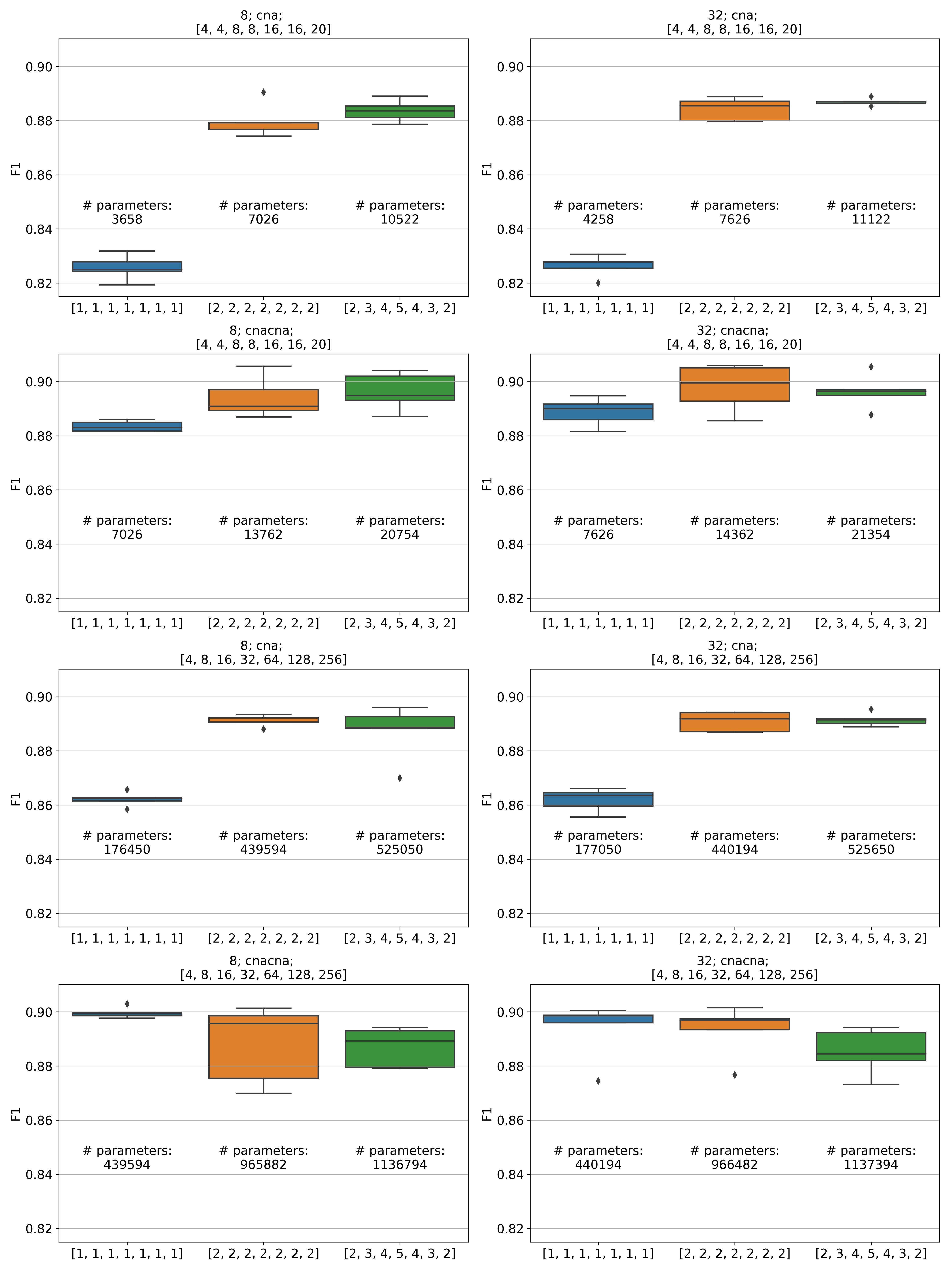}
\end{center}
\caption{F1 metric of model performance on the number of ResNet blocks in a group.}
\label{fig:blocks}
\end{figure*}

Same pattern appears on Fig~\ref{fig:blocks}: while overall number of parameters in the model does not exceed $5\times10^5$, more block seem to improve results of the model and, as the model size grows, this effect vanishes. 

A model of about $4\times10^5$ parameters with  \{[1,\;1,\;1,\;1,\;1,\;1,\;1,\;]\} blocks and \{[4,\;8,\; 16,\; 32,\; 64,\; 128,\; 256]\} filters outperforms larger models; nevertheless, it shows almost the same results as the models with more blocks and \{[4,\; 4,\; 8, \;8, \;16, \;16, \;20]\} filters, which have only about $20\times10^3$ parameters.

It is interesting to take a look at models with configurations like this:
\begin{align*}
	\{x; \; cna; \; [x]; \;[2,\; 2,\; 2,\; 2,\; 2,\; 2,\; 2]\}
\end{align*}
and
\begin{align*}
	\{x; \; cnacna; \; [x]; \;[1,\; 1,\; 1,\; 1,\; 1,\; 1,\; 1]\}
\end{align*}
as they only differ in a way skip connections are made. It seem that models with two convolutions between skip connection perform slightly better. It is reasonable since the skip connection represents identity operation and the whole ResNet block has more expressive power when the other branch is able to perform more complex transformations.

\section{Further research}
There is still a lot of aspects that require more deep and broad investigation. 

First, it is essential to extend the list of architectures under consideration, including VGG and MobileNet. In this work we have shown that models with moderate number of parameters ($10^4$-$10^6$) achieve the best results. It is interesting if the same pattern will be observed for other types of architectures, will the optimal range be shifted, etc.
Second, research grid should include different optimizers and their parameters. In this work we used Adam with default settings, but it is quite possible that models with different number of parameters experience better convergence with different optimizers and more sophisticated settings.
Finally, it is important to look deeper into dispersion of dispersions of scores that different configurations produce. When we observe a configuration with very similar scores in every repetition - is it a really stable configuration, or it is just a coincidence and we just need more repeats to observe its instability?

Those aspects, considered altogether in a single wide-scale experimental design might result in a comprehensive representation of atrial fibrillation classification task from the deep learning perspective.

\section{Conclusion}
Application of deep learning techniques to the real-world problems require comprehensive and systematic approach. This is the only way to draw meaningful conclusions from the conducted research. It means that apart from SOTA results we need something more - an understanding, why some kind of methods have led to SOTA in this task or this class of tasks.

As we have shown, different aspects of architecture configuration play significant role in overall performance of the model by itself and in combination with other parameters. We have found a feasible interval of number of parameters, in which almost all configurations show comparable performance, close to the best among all the models considered. The differences between models performance in this interval seem to be conditioned by parameters of configuration, such as layout and skip connection design. Some of these parameters have values that constantly produce better results, and it is reasonable to optimize these parameters in the first place.

Nevertheless, there is still a lot of research to be conducted in this direction to obtain a comprehensive understanding of the problem.

\clearpage
{\small
\bibliographystyle{ieee}
\bibliography{mybib}
}

\clearpage
\appendix

\setcounter{table}{0}
\renewcommand{\thetable}{\Alph{section}\arabic{table}}

\section{Tables}\label{tables}

\begin{table}[!ht]
\begin{center}
\def\arraystretch{1.1}
\begin{tabular}{| c | c | c | c | c |}
\hline
  & Model Configuration & \# Parameters & F1 Median & F1 STD
\\ \hline
1 & 8; cna; [4, 4, 8, 8, 16, 16, 20]; [1, 1, 1, 1, 1, 1, 1] & 3658 & 0.826 & 0.004
\\ \hline
2 & 32; cna; [4, 4, 8, 8, 16, 16, 20]; [1, 1, 1, 1, 1, 1, 1] & 4258 & 0.826 & 0.004
\\ \hline
3 & 8; cnacna; [4, 4, 8, 8, 16, 16, 20]; [1, 1, 1, 1, 1, 1, 1] & 7026 & 0.884 & 0.002
\\ \hline
4 & 8; cna; [4, 4, 8, 8, 16, 16, 20]; [2, 2, 2, 2, 2, 2, 2] & 7026 & 0.879 & 0.006
\\ \hline
5 & 32; cnacna; [4, 4, 8, 8, 16, 16, 20]; [1, 1, 1, 1, 1, 1, 1] & 7626 & 0.889 & 0.005
\\ \hline
6 & 32; cna; [4, 4, 8, 8, 16, 16, 20]; [2, 2, 2, 2, 2, 2, 2] & 7626 & 0.884 & 0.004
\\ \hline
7 & 8; cna; [4, 4, 8, 8, 16, 16, 20]; [2, 3, 4, 5, 4, 3, 2] & 10522 & 0.884 & 0.004
\\ \hline
8 & 32; cna; [4, 4, 8, 8, 16, 16, 20]; [2, 3, 4, 5, 4, 3, 2] & 11122 & 0.887 & 0.001
\\ \hline
9 & 8; cnacna; [4, 4, 8, 8, 16, 16, 20]; [2, 2, 2, 2, 2, 2, 2] & 13762 & 0.894 & 0.007
\\ \hline
10 & 32; cnacna; [4, 4, 8, 8, 16, 16, 20]; [2, 2, 2, 2, 2, 2, 2] & 14362 & 0.898 & 0.008
\\ \hline
11 & 8; cnacna; [4, 4, 8, 8, 16, 16, 20]; [2, 3, 4, 5, 4, 3, 2] & 20754 & 0.896 & 0.006
\\ \hline
12 & 32; cnacna; [4, 4, 8, 8, 16, 16, 20]; [2, 3, 4, 5, 4, 3, 2] & 21354 & 0.896 & 0.006
\\ \hline
13 & 32; cnacna; [4, 8, 12, 20, 32, 52, 84]; [1, 1, 1, 1, 1, 1, 1] & 64202 & 0.901 & 0.005
\\ \hline
14 & 32; ncnacn; [4, 8, 12, 20, 32, 52, 84]; [1, 1, 1, 1, 1, 1, 1] & 64522 & 0.892 & 0.008
\\ \hline
15 & 32; cnacna; [4, 8, 12, 20, 32, 52, 84]; [2, 3, 4, 5, 4, 3, 2] & 172154 & 0.891 & 0.007
\\ \hline
16 & 32; ncnacn; [4, 8, 12, 20, 32, 52, 84]; [2, 3, 4, 5, 4, 3, 2] & 173314 & 0.896 & 0.007
\\ \hline
17 & 8; cna; [4, 8, 16, 32, 64, 128, 256]; [1, 1, 1, 1, 1, 1, 1] & 176450 & 0.862 & 0.002
\\ \hline
18 & 32; cna; [4, 8, 16, 32, 64, 128, 256]; [1, 1, 1, 1, 1, 1, 1] & 177050 & 0.862 & 0.004
\\ \hline
19 & 8; cna; [4, 8, 16, 32, 64, 128, 256]; [2, 2, 2, 2, 2, 2, 2] & 439594 & 0.891 & 0.002
\\ \hline
20 & 8; cnacna; [4, 8, 16, 32, 64, 128, 256]; [1, 1, 1, 1, 1, 1, 1] & 439594 & 0.899 & 0.002
\\ \hline
21 & 32; cna; [4, 8, 16, 32, 64, 128, 256]; [2, 2, 2, 2, 2, 2, 2] & 440194 & 0.891 & 0.003
\\ \hline
22 & 32; cnacna; [4, 8, 16, 32, 64, 128, 256]; [1, 1, 1, 1, 1, 1, 1] & 440194 & 0.894 & 0.010
\\ \hline
23 & 8; cna; [4, 8, 16, 32, 64, 128, 256]; [2, 3, 4, 5, 4, 3, 2] & 525050 & 0.887 & 0.009
\\ \hline
24 & 32; cna; [4, 8, 16, 32, 64, 128, 256]; [2, 3, 4, 5, 4, 3, 2] & 525650 & 0.892 & 0.002
\\ \hline
25 & 8; cnacna; [4, 8, 16, 32, 64, 128, 256]; [2, 2, 2, 2, 2, 2, 2] & 965882 & 0.888 & 0.013
\\ \hline
26 & 32; cnacna; [4, 8, 16, 32, 64, 128, 256]; [2, 2, 2, 2, 2, 2, 2] & 966482 & 0.893 & 0.009
\\ \hline
27 & 8; cnacna; [4, 8, 16, 32, 64, 128, 256]; [2, 3, 4, 5, 4, 3, 2] & 1136794 & 0.887 & 0.007
\\ \hline
28 & 32; cnacna; [4, 8, 16, 32, 64, 128, 256]; [2, 3, 4, 5, 4, 3, 2] & 1137394 & 0.885 & 0.008
\\ \hline
29 & ResNet18 & 3843138 & 0.844 & 0.002
\\ \hline
30 & ResNet34 & 7217474 & 0.853 & 0.007
\\ \hline
\end{tabular}
\end{center}
\caption{Configuration, parameters and results for tested ResNet architecture variations.}
\label{tab:architectures}
\end{table}
\end{document}